\documentclass[sigconf]{acmart}

\AtBeginDocument{%
  }

\usepackage{bbm}
\usepackage{algorithm}
\usepackage{xcolor,colortbl}
\usepackage{balance}
\usepackage{nicematrix}

\usepackage{algorithmic}

\setcopyright{acmlicensed}
\copyrightyear{2024}
\acmYear{2024}

\acmISBN{978-1-4503-XXXX-X/18/06}

\copyrightyear{2024}
\acmYear{2024}
\setcopyright{acmlicensed}
\acmConference[CIKM '24] {Proceedings of the 33rd ACM International Conference on Information and Knowledge Management}{October 21--25, 2024}{Boise, ID, USA.}
\acmBooktitle{Proceedings of the 33rd ACM International Conference on Information and Knowledge Management (CIKM '24), October 21--25, 2024, Boise, ID, USA}
\acmISBN{979-8-4007-0436-9/24/10}
\acmDOI{10.1145/3627673.3679947}

\begin{document}

\title{SurvReLU: Inherently Interpretable Survival Analysis via Deep ReLU Networks}


\author{Xiaotong Sun}
\orcid{0009-0004-3520-4369}
\affiliation{%
  \institution{University of Arkansas}
  \city{Fayetteville}
  \state{AR}
  \country{USA}}
\email{xs018@uark.edu}

\author{Peijie Qiu}
\orcid{0000-0002-1591-5436}
\affiliation{%
  \institution{Washington University in St. Louis}
  \city{St. Louis}
  \state{MO}
  \country{USA}}
\email{peijie.qiu@wustl.edu}

\author{Shengfan Zhang}
\orcid{0000-0001-9866-7177}
\affiliation{%
  \institution{University of Arkansas}
  \city{Fayetteville}
  \state{AR}
  \country{USA}}
\email{shengfan@uark.edu}

\renewcommand{\shortauthors}{Sun et al.}

\begin{abstract}
Survival analysis models time-to-event distributions with censorship. Recently, deep survival models using neural networks have dominated due to their representational power and state-of-the-art performance. However, their "black-box" nature hinders interpretability, which is crucial in real-world applications. In contrast, "white-box" tree-based survival models offer better interpretability but struggle to converge to global optima due to greedy expansion.
\emph{In this paper, we bridge the gap between previous deep survival models and traditional tree-based survival models through deep rectified linear unit (ReLU) networks.} We show that a deliberately constructed deep ReLU network (termed SurvReLU) can harness the interpretability of tree-based structures with the representational power of deep survival models. 
Empirical studies on both simulated and real survival benchmark datasets showed the effectiveness of the proposed SurvReLU in terms of performance and interoperability. The code is available at~\href{https://github.com/xs018/SurvReLU}{\color{magenta}{ https://github.com/xs018/SurvReLU}}.
\end{abstract}




\maketitle

\section{Introduction}
Survival analysis is a prominent area of interest in statistics. Unlike time series analysis, which identifies patterns at regular time intervals~\cite{lim2021time,chen2024timemil,cao2018brits}, survival analysis models the time-to-event distribution. 
Time-to-event data involves censorship due to incomplete follow-up records~\cite{lee2003statistical}, making the true time-to-event distribution unknown.
In the early stages, survival analysis is approached from traditional statistics. The non-parametric Kaplan-Meier estimator~\cite{kaplan1958nonparametric} models the time-to-event data with counting statistics but ignores time-dependent covariates. Cox proportional hazards (CPH) model~\cite{cox1992regression} addresses this limitation by proposing a semi-parametric model that approximates the true hazard rate by combining a baseline hazard rate with a linear combination of covariates. However, this linear relationship between the risk function and covariates is easily violated in real scenarios. 

To alleviate this linear constraint, deep neural networks are introduced to statistical survival models~\cite{faraggi1995neural}, due to their power to approximate arbitrary functions. Following this vein, DeepSurv~\cite{katzman2018deepsurv} replaces the linear model in CPH with a multi-layer perceptron (MLP) while retaining the CPH framework. To improve scalability and reduce computational overhead in DeepSurv, \cite{kvamme2019time} proposes a case-control approximation to the Cox partial likelihood, which better leverages stochastic optimization in deep learning. 
Beyond deep CPH models, assumption-free deep survival models have also been explored. In particular, \cite{lee2018deephit} directly learns the distribution of survival times using a deep neural network by jointly maximizing the likelihood of the first hitting time and the corresponding event while minimizing a concordance-like ranking loss. To further capture dependencies between time slots, a deep recurrent neural network is proposed to 
model the cumulative risk function~\cite{ren2019deep}.  In parallel,~\cite{chapfuwa2018adversarial} proposes an adversarial learning scheme to directly learn the non-parametric time-to-event distribution. Despite achieving state-of-the-art performance, these deep survival models remain challenging to interpret due to their "black-box" nature.

Tree-based survival models~\cite{gordon1985tree,kogalur2008random} are competing approaches to deep survival models. These models either minimize local homogeneity~\cite{gordon1985tree} (measured by log-rank test) or maximize local heterogeneity~\cite{wang2016functional} (measured by Kullback–Leibler divergence) between branches at each splitting node. Oblique survival trees \cite{kretowska2019oblique,bertsimas2022optimal,huisman2024optimal,zhang2024optimal}, which use non-axis-aligned splitting rules as opposed to axis-aligned survival trees~\cite{gordon1985tree}, have also been explored.~\cite{kogalur2008random,jaeger2019oblique} extend the concepts of survival trees to random (oblique) survival forests to enhance their generalizability. Although these "white-box" tree-based survival models offer more interpretability compared to deep survival models, they are typically shown to be inferior to deep survival models~\cite{katzman2018deepsurv,kvamme2019time,lee2018deephit,ren2019deep}. 
This may be due to their inability to guarantee convergence to global optima because of their greedy expansion and reliance on predefined splitting rules.


\begin{figure*}[!t]
    \centering
    \includegraphics[width=0.75\textwidth]{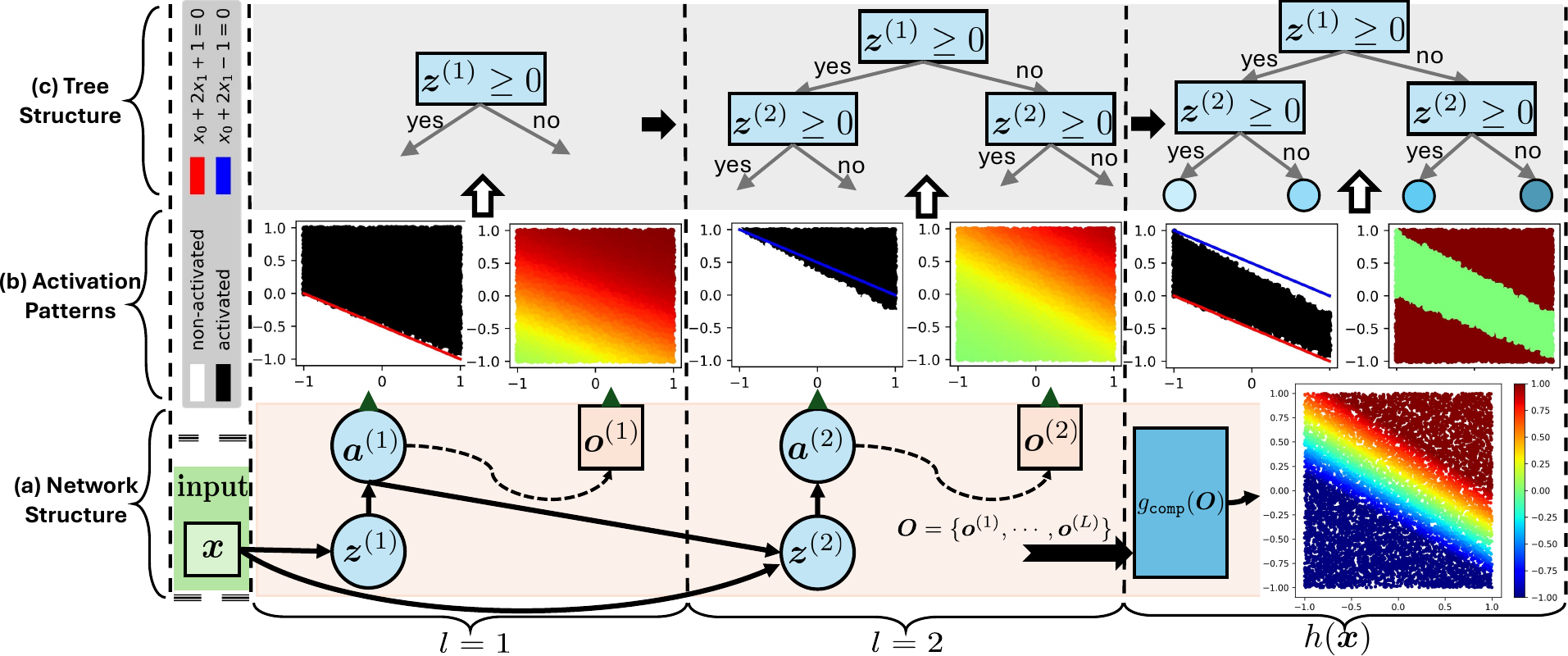}
    \vspace{-0.35cm}
    \caption{Illustrative example to approximate a linear hazard function: $h(\boldsymbol{x})=x_0 + 2 x_1$: (a) the network structure of the proposed SurvReLU network, (b) the decision boundary at each partitioning/layer, and (c) the resulting tree structures, respectively.}
    \label{workflow}
\end{figure*}

In this paper, we show that a thoughtfully designed ReLU network can achieve interpretability, by forming a tree structure, and maintain the representational power of a neural network. Specifically, we establish an explicit connection between the ReLU network and previous tree-based survival models, leveraging ReLU networks' ability to partition the input space into locally homogeneous regions like a tree (Sec.~\ref{sec:2.1}).
Second, we introduce a novel statistically-driven method to dynamically optimize the topology of the ReLU network specifically for survival analysis (Sec.~\ref{sec:2.2}). 
This can help compact the tree structure and improve the interpretability. To this end, we construct the novel ReLU network that is inherently interpretable like a tree (termed SurvReLU). Third, we show that SurvReLU can be optimized end-to-end with a flexible choice of loss functions (Sec.~\ref{sec:2.3}).
Experiments on both simulated and real datasets demonstrated the effectiveness of the proposed SurvReLU in terms of both performance and interpretability.

\section{Method}
The right-censored survival data is given in the form of $N$ triples $\{\boldsymbol{x}, T, E\}_{n=1}^{N}$, where $\boldsymbol{x}, T, E$ denote the $d-$dimensional covariates, the time of record, and an event indicator, respectively. With the random variable $t$ representing time, the objective is to model the probability of the event occurring over time $t$. This is commonly formulated to compute the survival function $S(t)$, i.e., the probability that an event occurs beyond time $t$:
\begin{equation}
    S(t) = \mathbb{P}(T>t) = 1 - \int_{0}^{t} f(s) \ ds,
\end{equation}
where $T$ is the true event time, and $f(t)$ denotes the probability density function. Alternatively, the hazard rate:
\begin{equation}
    h(t) = \frac{f(t)}{S(t)} = \lim_{\Delta t \rightarrow 0} \frac{\mathbb{P}(t \leq T < t + \Delta t | T \geq t)}{\Delta t},
\end{equation}
which specifies the risk at time $t$, is rather commonly used in most survival models~\cite{cox1992regression,katzman2018deepsurv,kvamme2019time,lee2018deephit,ren2019deep,chapfuwa2018adversarial,gordon1985tree,kogalur2008random}. We also follow the same convention to predict hazard rate in this paper. 
Following this definition, most previous survival models can be abstracted as $h_{\boldsymbol{x}}(t) = g_{\theta}(\boldsymbol{x}, t)$, where $g(\cdot)$ is a function parameterized by $\theta$ that maps covariates $\boldsymbol{x}$ to the corresponding hazard rate. Accordingly, the function $g(\cdot)$ can be a linear model~\cite{cox1992regression}, nonlinear neural networks~\cite{katzman2018deepsurv,kvamme2019time,lee2018deephit,ren2019deep}, and non-parametric survival trees~\cite{gordon1985tree,wang2016functional,jaeger2019oblique}. 

\emph{The key purpose of this paper is to bridge the gap between previous deep survival models and survival trees via a ReLU network.} 
Similar to tree-based models, ReLU networks can partition input space into locally homogeneous and disjoint polyhedrons~\cite{montufar2014number,lee2020oblique,lee2018towards}. Instead of using predefined partitioning rules in previous survival trees, the ReLU network learns partitioning rules from input data. This empowers the ReLU network to have the same representational power as neural networks, i.e., approximating arbitrary functions~\cite{montufar2014number}. 

\subsection{Network Architecture}\label{sec:2.1}
A deep ReLU network is an MLP with ReLU activation functions. Given an $L$-layer ReLU network, the output from the $l$-th layer with $l \in \{1, \cdots, L\}$  can be expressed as 
\begin{equation}\label{eq:3}
\begin{split}
     \boldsymbol{a}^{(l)} = \sigma(\boldsymbol{z}^{(l)}), \ \boldsymbol{z}^{(l)} = \boldsymbol{W}^{(l)} \boldsymbol{a}^{(l-1)} + \boldsymbol{b}^{(l)} , \
     \boldsymbol{a}^{(0)} = \textbf{x},
\end{split}
\end{equation}
where $\sigma(\cdot)$ denotes the activation function, $\boldsymbol{W}^{(l)}$ and $\boldsymbol{b}^{(l)}$ are the weight matrix and bias vector, respectively. 

The end-to-end behavior of the ReLU network is uniquely determined by its activation pattern~\cite{montufar2014number,lee2020oblique,lee2018towards}, i.e., the derivative of the ReLU activation function w.r.t. its input $\boldsymbol{o}^{(l)} = \partial \boldsymbol{a}^{(l)} / \partial \boldsymbol{z}^{(l)}$. Once the activation patterns $\boldsymbol{O} = \{\boldsymbol{o}^{(1)}, \cdots, \boldsymbol{o}^{(L)}\}$ of the ReLU network are fixed, the partitioning of the input space is uniquely determined, as $\boldsymbol{o}^{(l)}$ is locally constant (see~\cite{lee2020oblique}). Since any function taking locally constant input remains locally constant, we can use a composite function $g_{\texttt{comp}}$ to map the activation patterns to the hazard rate $g_{\texttt{comp}}: \boldsymbol{O} \rightarrow h(\boldsymbol{x})$, where $g_{\texttt{comp}}$ can be any models (e.g., linear model and MLP).
As a result, the activation patterns $\boldsymbol{O}$ and composite function $g_{\texttt{comp}}$ in the ReLU network serve as the nodes and the leaves in survival trees, respectively. Unlike the previous tree-based survival models where the leaf node is restricted to either Nelson–Aalen estimator~\cite{kogalur2008random} or CPH model~\cite{jaeger2019oblique}, the learnable composite function enables integration of more advanced loss functions (e.g., ranking loss~\cite{lee2018deephit}; see discussion in Sec.~\ref{sec:2.3}).

One notable difference between survival trees~\cite{gordon1985tree,wang2016functional,jaeger2019oblique} and the ReLU network is the dependence of input $\boldsymbol{x}$ at each partitioning/layer. To incorporate $\boldsymbol{x}$ at each layer, Eq.~(\ref{eq:3}) is modified as 
\begin{equation}
    \boldsymbol{z}^{(l)} = \boldsymbol{W}^{(l)} \texttt{concat}[\boldsymbol{x}; \boldsymbol{a}^{(l-1)}] + \boldsymbol{b}^{(l)}, \ \forall l > 2,
\end{equation}
where $\texttt{concat}[\cdot]$ denotes concatenation operation. The resulting network architecture is shown in Fig.~\ref{workflow}(a). This modification enables the partitioning of polyhedrons at each layer to resemble a non-axis-aligned tree-splitting rule, where the absolute value of each entry in $\boldsymbol{W}^{(l)}$ indicates the importance of each covariate at the $l$-th layer. Note that a sparse weight matrix $\boldsymbol{W}^{(l)}$ (i.e., $\boldsymbol{W}^{(l)}$ has only one nonzero entry) can also lead to an axis-aligned splitting rule.

\emph{Up to now, we have established the explicit connection between the ReLU network and previous axis-aligned~\cite{gordon1985tree,wang2016functional} and non-axis-aligned~\cite{jaeger2019oblique} survival trees. Since the derived ReLU network itself is a neural network, its connection to other deep survival models~\cite{katzman2018deepsurv,kvamme2019time,lee2018deephit,ren2019deep} is trivial.}

\begin{figure}[!t]
    \centering
    \includegraphics[width=0.45\textwidth]{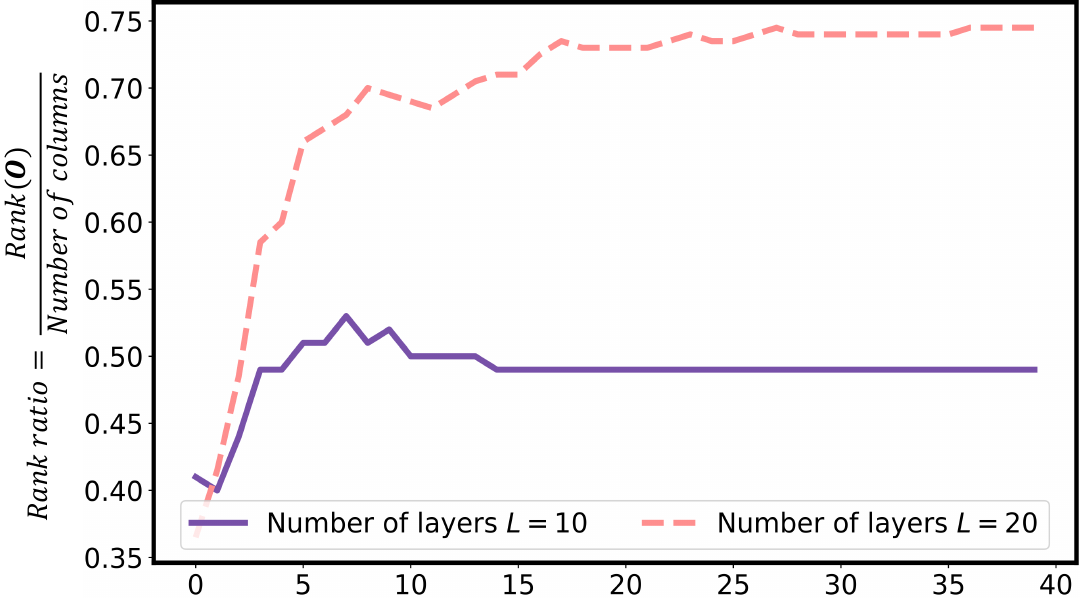}
    \vspace{-0.35cm}
    \caption{The change of rank ratio of matrix $\boldsymbol{O}$ over epochs.}
    \label{fig:rank}
    \vspace{-0.2cm}
\end{figure}


\subsection{Statistically-Driven Topology Optimization}\label{sec:2.2}
Although we have constructed a ReLU network whose behavior inherently resembles a tree, it is not specifically optimized for survival analysis. One notable limitation of the previously derived ReLU network in Sec.~\ref{sec:2.1} is that its network topology is defined a priori. This implies that the tree constructed by the ReLU network is fully expanded (i.e., a fully-expanded dense tree). 

As the activation patterns $\boldsymbol{O}$ uniquely encoded the partitioning of input space in the derived ReLU network, the topology of the learned tree structure can be reflected by $\boldsymbol{O}$. We observe that the activation patterns matrix $\boldsymbol{O}$ is typically low-ranked (see Fig.~\ref{fig:rank}). This suggests that the learned ReLU network, however, produces a sparse tree topology: some partitions/nodes should be pruned. Based on this observation, we employ the log-rank test to localize which node should be pruned. Specifically, at each forward pass, we can reconstruct the tree structure of the derived ReLU network recursively (see Fig.~\ref{workflow}(c)). Simultaneously, we can perform the log-rank test at each splitting node and merge the nodes whose p-value from the log-rank test is less than 0.05 ($p < 0.05$) by setting the corresponding activation patterns to 0. This enables us to dynamically optimize the derived ReLU network's topology and automatically prune the resulting tree structures.

However, when to perform the proposed topology optimization remains an open challenge. Model performance may deteriorate when optimizing network topology after training~\cite{molchanov2016pruning}. However, optimizing network topology at early training iterations may restrict its representational capacity~\cite{blalock2020state}. Again, we can leverage the rank of the activation patterns $\boldsymbol{O}$ to decide when to optimize the network topology. 
To achieve this, we keep track of the changes in the rank of the matrix $\boldsymbol{O}$ over time. Once the topology of the learned tree structures is unchanged, we optimize the network topology.

\emph{To this end, we have constructed a ReLU network that is inherently interpretable like a survival tree but has a topology that can be learned and dynamically optimized. We term this novel ReLU network as SurvReLU, as it is tailored for survival analysis.} 

\subsection{Model Learning}\label{sec:2.3}
The proposed SurvReLU can be optimized in an end-to-end fashion. A summary of training a single iteration of SurvReLU is specified in Algorithm~\ref{alg:1}. It is noteworthy that the optimization of network topology (line 4-6 in Algorithm~\ref{alg:1}) is only performed when the rank of matrix $\boldsymbol{O}^{(l)}$ is unchanged as discussed in Sec.~\ref{sec:2.2}. 

\noindent \textbf{Loss function.} The end-to-end parameterization of SurvReLU enables a flexible choice of the loss function. We consider both continuous-time~\cite{katzman2018deepsurv} and discrete-time~\cite{lee2018deephit} survival losses in this paper due to their popularity and success; while other choices can be alternatives (e.g., adversarial loss~\cite{chapfuwa2018adversarial}). The resulting models are named SurvReLU (Cont.) and SurvReLU (Disc.), respectively. 

\noindent \textbf{Complexity analysis.} Compared to standard MLP, SurvReLU computes the derivative of each neuron w.r.t. $\boldsymbol{x}$, which requires an additional back-propagation ($\mathcal{O}(L^3)$). By leveraging the dynamic programming trick as outlined in~\cite{lee2020oblique}, this can be reduced to $\mathcal{O}(L^2)$. The log-rank test also introduces additional computation, which will take $\mathcal{O}(N\log N)$ to sort the time $T$. However, this can be handled once at the very beginning of each training iteration.  

\begin{algorithm}[!t]
\caption{SurvReLU Training (single iteration)} 
\begin{algorithmic}[1]\label{alg:1}

\REQUIRE Batch data: $\{\boldsymbol{x}, T, E\}_{n=1}^{N}$, ReLU network: 
 [$g_{\theta}$; $g_{\texttt{comp}}$]
\FOR{$l=1:L$}
\STATE $\boldsymbol{z}^{(l)} = \boldsymbol{W}^{(l)} \texttt{concat}[\boldsymbol{x}_n; \boldsymbol{a}^{(l-1)}] + \boldsymbol{b}^{(l)}$; $\boldsymbol{a}^{(l)} = \sigma(\boldsymbol{z}^{(l)})$
\STATE compute activation patterns $o_n^{(l)}$ at $l$-th layer
\STATE reconstruct the tree node at the $l$-th layer (see Fig.~\ref{workflow}(c))
\STATE log-rank test on two branches ($\boldsymbol{z^{(l)}} \geq 0$ and $\boldsymbol{z^{(l)}} < 0$)
\STATE update the tree structure topology (as specified in Sec.~\ref{sec:2.2})

\ENDFOR
\STATE $h(\boldsymbol{x}_n) = g_{\texttt{comp}}(\{o^{(1)}_n, \cdots, o^{(L)}_n\})$
\STATE compute loss and update network parameters by SGD
\end{algorithmic}
\end{algorithm}

\begin{table}[!t]
\vspace{-0.2cm}
  \caption{Performance comparison using $C^{td}_{\tau} [P(C > \tau) = 10^{-8}]$.}
  \vspace{-0.3cm}
  \centering
  \resizebox{0.49\textwidth}{!}{
  \begin{NiceTabular}{l|c|c|c|c}[colortbl-like]
    \toprule
    Methods  & Simulated Linear & Simulated Gaussian  &  SUPPORT & METABRIC  \\
    \hline
    CPH~\cite{cox1992regression} & 0.774 (0.772, 0.775) & 0.507 (0.505, 0.509) & 0.569 $\pm$ 0.010 &   0.636 $\pm$ 0.012  \\
    DeepSurv~\cite{katzman2018deepsurv} & 0.774 (0.772, 0.776) & 0.649 (0.647, 0.651) & 0.611 $\pm$ 0.007  &  0.652 $\pm$ 0.012   \\
    Cox-CC~\cite{kvamme2019time} & 0.775 (0.757, 0.789) & 0.649 (0.627, 0.672) & 0.613 $\pm$ 0.003 &  0.649 $\pm$ 0.009   \\
    DeepHit~\cite{lee2018deephit} & \underline{0.777 (0.760,  0.795)} & \underline{0.658  (0.637, 0.679)} & \underline{0.641 $\pm$  0.004} &  \underline{0.677 $\pm$ 0.016} \\ 
    \midrule
    ST~\cite{gordon1985tree} & 0.694 (0.676, 0.715) & 0.571 (0.544, 0.591)  &  0.554 $\pm$ 0.007  & 0.563 $\pm$ 0.014  \\
    OST~\cite{jaeger2019oblique} & 0.681 (0.658, 0.701) & 0.541  (0.518, 0.562) & 0.542 $\pm$  0.005 &  0.556 $\pm$ 0.013 \\
     RSF~\cite{kogalur2008random} & 0.765 (0.763, 0.766) & 0.646 (0.643, 0.648) & 0.614 $\pm$ 0.004  &  0.643 $\pm$ 0.011   \\
     ORSF~\cite{jaeger2019oblique} & 0.774 (0.758, 0.794) & 0.631 (0.605, 0.649) & 0.636 $\pm$ 0.006 & 0.674 $\pm$ 0.014\\
    \midrule
    \midrule
    \rowcolor{pink!30}
    SurvReLU (Cont.) & \textbf{0.780 (0.768, 0.795)} & \textbf{0.660 (0.636, 0.678)}  & 0.620 $\pm$ 0.005 &  0.659 $\pm$ 0.009   \\
    \rowcolor{pink!30}
    SurvReLU (Disc.) & \textbf{0.780 (0.764, 0.797)} & 0.659  (0.639,  0.682)  & \textbf{0.645 $\pm$ 0.005} &  \textbf{0.679 $\pm$ 0.017}   \\
    \bottomrule
  \end{NiceTabular}
  }
  \label{tab:all_results}
  \vspace{-0.2cm}
\end{table}
\begin{figure*}[!t]
    \centering
    \includegraphics[width=0.95\textwidth]{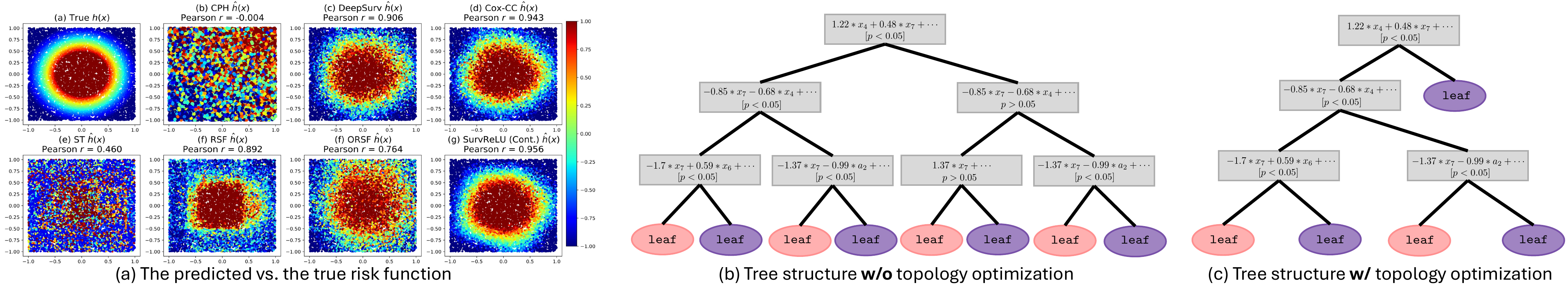}
    \vspace{-0.3cm}
    \caption{(a) The predicted risk function versus the true Gaussian risk function on the simulated Gaussian dataset. (b) and (c): The resulting tree structures of SurvReLU w/ and w/o the proposed topology optimization in Sec.~\ref{sec:2.2} on METABRIC dataset.}
    \label{fig:viz}
\end{figure*}

\section{Experiments}
We validated SurvReLU on two simulated and two real datasets.
\noindent \textbf{Datasets.}
The simulated datasets were sourced from~\cite{katzman2018deepsurv}, consisting of a training ($N = 4000$), validation ($N = 1000$), and testing ($N = 1000$) set. Each sample includes $d = 10$ covariates (i.e., $x_0, x_1, \cdots, x_9$) drawn from a uniform distribution on $[-1,1)$. The event time $T$ was generated based on an exponential Cox model. Further information about the simulation process can be found in~\cite{katzman2018deepsurv}. These simulated datasets encompass two subcategories characterized by either a linear risk function (i.e., $x_1$: $h(x) = x_0 + 2x_1$) or a Gaussian risk function (i.e., $h(\boldsymbol{x}) = \log(r_{max}) \exp(-(x_0^2+x_1^2)/2\sigma^2)$; $r_{max}=5.0$ and $\sigma = 0.5$).
Additionally, two benchmark real datasets with censorship of $32\%$ and $42\%$ were used:
\textbf{(i)} SUPPORT dataset, originating from the Study to Understand Prognoses Preference Outcomes and Risks of Treatment~\cite{knaus1995support}. It comprises 8873 patients (patients with missing features were dropped) with 14 covariates. 
\textbf{(ii)} METABRIC dataset is designed to improve breast cancer treatment recommendations~\cite{curtis2012genomic}, including gene expression and clinical data from 1,980 patients. The data was processed to integrate four gene indicators and five clinical features, resulting in 9 covariates. 

\noindent \textbf{Evaluation metrics.} The evaluation of the performance for survival analysis was carried out by the time-dependent concordance index ($C^{td}_{\tau}$). Specifically, C-index is designed to assess a model's ability to predict the ordering of event times through ranking. 

\noindent \textbf{Experimental setup.}
\textbf{(i)} On the simulated datasets, we adhered to the default training, validation, and testing data splits outlined in~\cite{katzman2018deepsurv}. All the covariates were z-scored to have a zero mean and unit variance. The 95\% confidence intervals were obtained by bootstrapping~\cite{katzman2018deepsurv}. 
\textbf{(ii)} We conducted five-fold cross-validation for the SUPPORT and METABRIC datasets to thoroughly assess the performance and generalizability of all methods. The mean ($\pm$ standard deviation) of C-index is reported. For preprocessing, we applied one-hot encoding to all categorical variables and z-score for all numerical variables. 
\textbf{(iii)} The hyperparameter tuning for the proposed model (i.e., learning rate, batch size, and $L$) was carried out by the random search~\cite{bergstra2012random}. The optimal learning rate was found to be $0.1$. The batch size was set to 1024. We recommend a large batch size to ensure stable statistics in the log-rank test. The optimal number of layers was [6, 30, 16, 14] for simulated Linear, simulated Gaussian, SUPPORT, and METABRIC, respectively. All the experiments were implemented in \texttt{PyTorch} and performed on a Nvidia GTX 1080Ti.

\section{Results}
\noindent \textbf{Main results.} SurvReLU competed favorably with other deep survival and tree-based survival models on both simulated and real datasets in terms of $C^{td}$ (Table~\ref{tab:all_results}). We observed that \textbf{(i)} on the simulated datasets, SurvReLU (Cont.) and SurvReLU (Disc.) achieved similar performance but still outperformed other competing methods. Remarkably, SurvReLU approximated true Linear/Gaussian risk functions as effectively as other deep survival models (see Fig.~\ref{fig:viz}(a)). However, the axis-aligned and non-axis-aligned tree-based survival models showed inferior performance in modeling the risk function. This suggests that SurvReLU retains the representational capacity of a neural network. Although SurvReLU has a non-axis-aligned partitioning rule similar to ORSF, we conjecture that the superiority of SurvReLU over other tree-based survival models arises from its end-to-end optimization instead of greedy expansion (e.g., fitting a CPH at each split in ORSF). \textbf{(ii)} On the real datasets, SurvReLU (Cont.) outperformed other competing continuous-time deep survival models (DeepSurv and Cox-CC). SurvReLU (Disc.) outperformed discrete-time DeepHit and other tree-based survival models, even without model ensembles. Importantly, SurvReLU resulted in a compact tree structure that was inherently as interpretable as tree-based survival models (Fig.~\ref{fig:viz}(b) and (c)), a feature none of the other deep survival models can provide.

\begin{figure}[!t]
    \centering
    \includegraphics[width=0.472\textwidth]{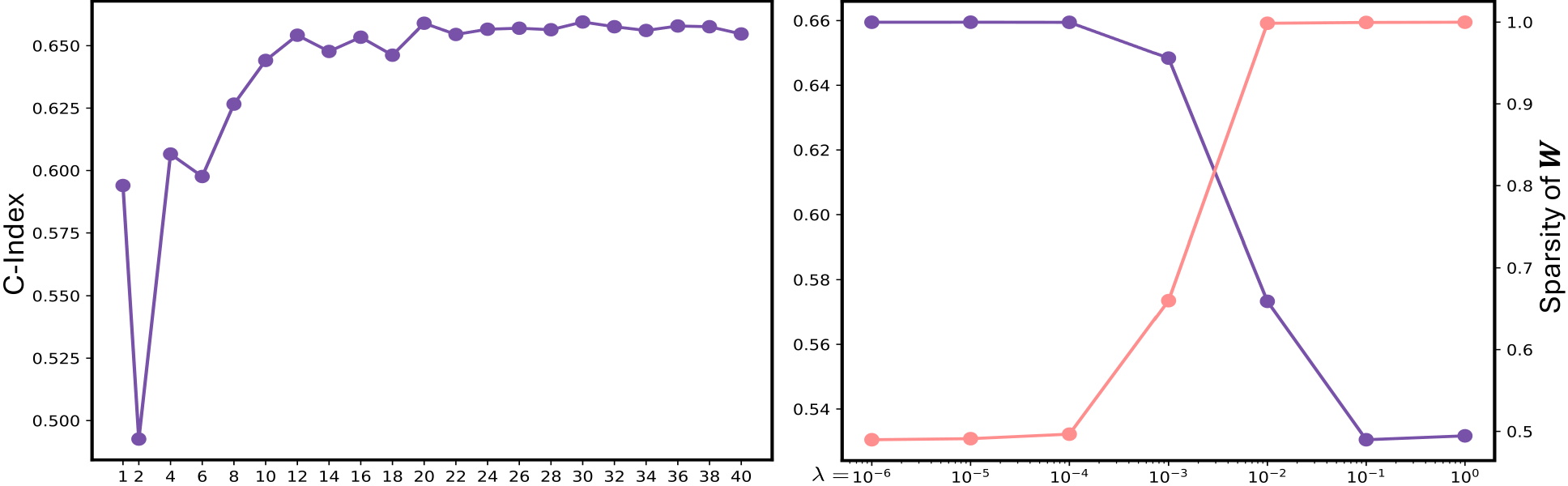}
    \vspace{-0.3cm}
    \caption{Ablations on number of layers $L$ (Left) and sparsity of $\boldsymbol{W}$ (Right) using the simulated Gaussian dataset.}
    \label{fig:ablation}
\end{figure}

\noindent \textbf{Ablation studies.} We conducted ablation on the number of layers $L$ in a SurvReLU network as well as the sparsity of the weight matrices $\boldsymbol{W} = \{\boldsymbol{W}^{(1)}, \cdots, \boldsymbol{W}^{(L)} \}$. The sparsity of $\boldsymbol{W}$ is enforced by recursively \texttt{soft-thresholding} $\boldsymbol{W}$ with a strength $\lambda$ (see e.g.~\cite{donoho1995noising,kusupati2020soft,xiao2023sc,qiu2023sc}).
We observed that \textbf{(i)} similar to many deep survival models, the performance of SurvReLU typically saturated after reaching a certain number of layers (e.g., $L=20$ in Fig.~\ref{fig:ablation}(\textbf{Left})). \textbf{(ii)} The performance of SurvReLU was traded-off with the sparsity of~$\boldsymbol{W}$ and hence the interpretability of the model (Fig.~\ref{fig:ablation}(\textbf{Right})). 
This is because a more sparse $\boldsymbol{W}$ pushes SurvReLU to result in an axis-aligned tree structure (see Sec.~\ref{sec:2.1}), which is better to interpret.


\section{Conclusion and Future Work}
In this paper, we bridge the gap between deep survival models and survival trees by proposing a statistically-driven ReLU network for survival analysis (SurvReLU). Experimental results on both simulated and real datasets showed that SurvReLU competed favorably with previous deep and tree-based survival models. Notably, it retains the interpretability of tree-based models without compromising performance. However, unlike other tree-based models, SurvReLU did not use model ensembles, which could be explored in future work to enhance its generalizability and performance.

\begin{acks}
This work was supported in part by a National Science Foundation grant (Award No: 1920920) and the Open Access Publishing Fund administered through the University of Arkansas Libraries. 
\end{acks}

\bibliographystyle{ACM-Reference-Format}
\balance
\bibliography{refs}


\end{document}